%% file: ms.tex
\documentclass{article}

\PassOptionsToPackage{numbers}{natbib}
\usepackage[final]{nips_2017}

\usepackage[utf8]{inputenc} % allow utf-8 input
\usepackage[T1]{fontenc}    % use 8-bit T1 fonts
\usepackage{hyperref}       % hyperlinks
\usepackage{url}            % simple URL typesetting
\usepackage{booktabs}       % professional-quality tables
\usepackage{amsfonts}       % blackboard math symbols
\usepackage{nicefrac}       % compact symbols for 1/2, etc.
\usepackage{microtype}      % microtypography
\usepackage{graphicx}
\usepackage{epstopdf}
\usepackage{subfig}
\usepackage{pgffor}
\usepackage{amsmath}
\usepackage{bmpsize}

\newif\ifpaper

\title{Observational Learning \\by
Reinforcement Learning}%: learning aided by observing another agent}

\author{
  Diana Borsa\\
  DeepMind\\
  \texttt{borsa@google.com} \\
   \And
  Bilal Piot\\
  DeepMind\\
  \texttt{piot@google.com} \\
   \AND
  Rémi Munos\\
  DeepMind\\
  \texttt{munos@google.com} \\
   \And
  Olivier Pietquin\\
  DeepMind\\
  \texttt{pietquin@google.com} \\
}

\begin{document}

\maketitle

\begin{abstract}

    Observational learning is a type of learning that occurs as a function of observing, retaining and possibly replicating or imitating the behaviour of another agent. 
    It is a core mechanism appearing in various instances of social learning and has been found to be employed in several intelligent species, including humans. In this paper, we investigate to what extent the explicit modelling of other agents is necessary to achieve observational learning through machine learning. Especially, we argue that observational learning can emerge from pure Reinforcement Learning (RL), potentially coupled with memory. Through simple scenarios, we demonstrate that an RL agent can leverage the information provided by the observations of an other agent performing a task in a shared  environment. The other agent is only observed through the effect of its actions on the environment and never explicitly modeled. Two key aspects are borrowed from observational learning: i) the observer behaviour needs to change as a result of viewing a 'teacher' (another agent) and ii) the observer needs to be motivated somehow to engage in making use of the other agent's behaviour. The later is naturally modeled by RL, by correlating the learning agent's reward with the teacher agent's behaviour.

\end{abstract}
\input{paper}

\bibliographystyle{unsrt}
\bibliography{bibliography}
%\input{appendix}

% paper only
\iffalse
% % appendix only
% \iffalse
\newpage
\input{appendix}
\fi

\end{document}

%% file: paper.tex
\section{Introduction}

Humans have evolved to live in societies and a major benefit of that is the ability to leverage the knowledge of parents, ancestries or peers to aid their understanding of the world and more rapidly develop skills deem crucial for survival. Most of this learning is done by observing the behaviour of the other agents; from this emerges role modeling, imitation or observational learning. We are particularly interested by observational learning in this paper and we define it as the ability for an agent to modify its behavior or to acquire information as an effect of observing another agent sharing its environment. 

In the machine learning literature, one of the most popular and successful ways for modeling goal-motivated learning agents is via Reinforcement Learning (RL)~\cite{sutton1998reinforcement,mnih2015human}. In the recent years, combining RL with the increased representational power of deep learning~\cite{lecun2015deep} and the memory capabilities of recurrent models (LSTMs/GRUs)~\cite{hochreiter1997long,chung2015gated} has lead to a string of impressive successes ranging from video-game playing~\cite{mnih2015human} to 3D navigation tasks~\cite{mnih2016asynchronous, mirowski2016learning} and robotics~\cite{levine2016end}.
Motivated in part by these, we here want to study if observational learning can naturally emerge in DeepRL agents empowered with memory. 
The main questions we would want to answer are then: \textit{is (deep) RL coupled with memory enough to successfully tackle observational learning?} Will the RL agent learn to ignore or leverage the teacher? Is the RL signal enough for the emergence of more complex behaviour like\textit{ imitation}, \textit{goal emulation} or \textit{information seeking}? In other words, we want to understand to whether other agents have to explicitly be modeled as such by learning agents or if the combination of perception (deep nets), memory (recurrent nets) and motivation (RL) is enough to learn from the sole observation of other agents' effects on a shared environment. 
%For this, we consider an RL agent that shares its environment with another agent (teacher) and we would like to explore if the RL agent (student) can take advantage of the sheer presence of this teacher to improve its policy or speed up its learning. The student has access to the teacher only through its first-person observations of the world and the learning will be powered solely through the external reward signal coming from the environment. 

It is worth noting that similar questions have been investigated in the cognitive and behaviour science community. In his work in 1977, Bandura proposed and coined the term 'observational learning' or social learning \cite{bandura1977social,bandura1963social}. According to him, observational learning differs from imitative learning in that it does not strictly require a duplication of the behavior exhibited by the teacher/expert. Heyes (1993) distinguished imitation and non-imitative social learning in the following way: imitation occurs when animals learn about behavior from observing conspecifics (sometimes other species), whereas non-imitative social learning occurs when animals learn about the environment from observing others (conspecifics or not)~\cite{heyes1993imitation}, meaning that one can learn about the dynamics of its environment only by observing other agents evolving in this environment. 

%[Need more here and background lit] 
Learning with the help of a teacher (or expert) is by no means a new idea in machine learning neither. Imitation learning has a long standing in the machine learning literature~\cite{schaal1999imitation,argall2009survey}. In this body of work, one can distinguish two major ideas: i) \textit{behaviour cloning}, where we are regressing directly onto the policy of another agent/expert~\cite{pomerleau1989alvinn,ratliff2007imitation}, or ii) \textit{inverse RL}, where we are trying to infer a reward function from the behaviour of other agents~\cite{russell1998learning,ng2000algorithms} and then use this, in conjunction with RL techniques, to optimize this inferred function. While these methods have been successfully applied to a variety of tasks \cite{pomerleau1989alvinn,atkeson1997robot,ziebart2008maximum,kolter2008hierarchical,ross2011reduction,ho2016showing,ho2016generative,piot2014predicting,neu2009training}, one problematic aspect of both these scenarios is that they almost always need to provide the learning agent with the expert trajectories in the same state-action space as the learner. Otherwise, some explicit mapping between the learner and the teacher state space has to be discovered~\cite{stadie2017third}. 
As previously argued/recognized in \cite{stadie2017third}, these can be somewhat restrictive and unrealistic requirements. 
%For instance, if our agent would want to learn from a human demonstrating a task or even a robot, it is non-trivial to map the observed behaviour into something our agent can readily use, employing the above computational techniques. 
Furthermore, in inverse RL one has to explicitly model the behaviour/trajectories coming from another agent and infer the reward signal. In order for this problem to become tractable, most of the time we need to make some structural assumptions about this reward -- like linearity in a given feature space, or smoothness~\cite{abbeel2004apprenticeship,piot2014boosted}. These assumptions might not model closely to the true reward signal and minor approximation errors can be easily amplified when planning onto this faulty signal~\cite{piotECML}. 

Given these increased complexities, we propose to study the simpler yet more natural alternative of observational learning, moving away from the traditional setup of learning from experts.
% -- inverse RL, behaviour cloning or, more generally, learning directly from expert data. 
%-- and we aim to investigate learning directly from observing another agent acting in a shared environment. 
%
This is not to say that we can address all of the problems tackled by the imitation learning literature. We are merely arguing that there might be scenarios where this level of modelling is not required and the RL agent can learn more directly through pure observations.
%
%We propose the following simple scenario: two agents in the same environment, at the same time. One agent is an expert -- knows how to do the task at hand, or knows the environment well enough to get around to potentially interesting, rewarding parts of the state space. Note that its behaviour needs not to be optimal in any of these. The second one, our learning agent (student), has impoverished knowledge of the environment/task at hand, and needs to learn how to perform the same or a similar task as the first agent (teacher). Please note that the two agents do not need to share the same task, but we assume there is information in the behaviour of the expert-agent that can aid the policy of the student. Nevertheless, for the sake of clarity, in this work we will focus mainly on agents who do share the same reward. This is a very natural setting that could potentially lead to imitation and/or apprenticeship learning. In this setup, we investigate if the student can benefit from the presence of the expert-agent. In particular, we want to do this without explicitly modelling the expert, nor artificially encouraging any interaction between the two agents.
% 
In this context, our main contribution is to exhibit scenarios where observational learning is emerging from a standard DeepRL algorithm (A3C~\cite{mnih2016asynchronous}) when combined or not with memory. In all our scenarios, the A3C agent (learner or student) shares its environment with another agent (expert or teacher) that has a better knowledge of the task to solve. The learner observes the expert through its sole perception of the environment. It is only rewarded for performing the task and doesn't receive any incentive to follow, imitate or interact with the expert. The expert is not aware that it is watched by the learner and is not meant to teach or provide extra information to the learner neither. By building tasks of increasing difficulty, we show that complex behaviours such as imitative and non-imitative learning emerge without explicit modeling of the expert. In addition, we provide some theoretical insights to explain why these behaviours are possible. 

In the next section we describe in more details our experimental design. Section~\ref{sec:formalism} provides the general background of RL and the theoretical foundations of this work. In Section~\ref{sec:exp} we provide the details of our experimental results before concluding in Section~\ref{sec:ccl}.

\section{Experimental design}

%In this work, we aim to investigate if a learning agent can benefit from the presence of an teacher in the environment and can learn to leverage this new information, without ever explicitly modelling this other agent. More interestingly we want to know if this behaviour could emerge through pure reinforcement learning. 
As explained in the introduction, we are primarily interested to see if an RL agent can learn to leverage the behaviour of an expert, based solely on 1) external reward (from the environment), ii) its ability to observe the consequences of the expert's actions in the environment. The learner does not have a notion of the agency of the teacher. This additional agent is simply part of the learner's environment and it can choose to ignore the presence of the expert if it deems this signal unimportant. It is worth noting that in all our case studies, the presence of the teacher does not impact the dynamics nor rewards of the RL agent. This is a necessarily assumption that will be formalized in Section \ref{sec:formalism}.

The first question we want to answer is whether the teacher's presence has any impact on the learner. For this we look at two scenarios: 1) the learner has perfect information and can learn an optimal behaviour on its own, 2) the learner has only partial information about the environment/task, but the expert's behaviour can provide additional information by providing a demonstration of the desired behaviour. In the first case, we do not expect a difference between the learnt policies with or without the expert. Nevertheless, when adding the teacher in the same environment, we are effectively expanding the state space of the learning agent. Thus, on top of the RL policy, now the student also needs to learn to ignore this extra signal in its state space.
In the second scenario however, the expert's behaviour contains crucial information for improving the student's policy. In this case, by ignoring the expert, the student can still complete the task at hand, but can only do so, sub-optimally. Now, there is a real incentive for the learning agent to pay attention to the teacher. Nevertheless, the student's own reward signal is still the one coming directly from the environment (which it would experience even without the other agent), but now our agent needs to somehow correlate this (potentially sparse) reward signal with the behaviour exhibited by the teacher. This is a highly non-trivial association the learning agent needs to make. And then learn to exploit it, in order to improve its policy.

If the two agents have the same reward structure, a good strategy for the learning agent would be to imitate, if possible, the expert's behaviour. This, in principle, is a much easier and safer policy than attempting to randomly explore the environment on its own. The student would only need to solve the local problem of following the expert, but would not need to worry about the global task - the global planning that is now done by the expert. Although this might not be optimal, this kind of behaviour is transferable between tasks and/or environments and could potentially provide the student with a better initial policy of exploring an unfamiliar environment. This could lead to a more principled/guided way to explore an unknown environment and have a major impact on the speed at which the agent discovers areas of interest, especially in a sparse reward setting.

Finally, we are interested in showing that the student can become autonomous and still perform the task optimally in the absence of the expert after learning from observations. Indeed, the final goal of a learning agent is to solve tasks on its own and it should still be able to reach that goal after having learned optimal policies from a teacher. 

% If the agents have different reward structure, the behaviour of the expert may still be informative for the other agent. (Last section)

\section{Learning by observing an expert}
\label{sec:formalism}
\subsection{Notation}
A Markov Decison Process (MDP) is a a tuple $\langle \mathcal{S}, \mathcal{A}, \mathcal{P}, \mathcal{R}, \gamma \rangle$ where $\mathcal{S}$ is a set of states, $\mathcal{A}$ is the set of actions available to the agent, $\mathcal{P}$ is the transitional kernel giving a probabily over next states given our current state and action, $\mathcal{R} : \mathcal{S} \times \mathcal{A} \rightarrow \mathbb{R}$ is a reward function and $\gamma \in[0,1]$ is a discount factor. 

We define a (stochastic) policy $\pi : \mathcal{S} \times \mathcal{A} \rightarrow [0,1]$ that maps states into a probability over the set of actions. Given such a policy $\pi$ we can define the value function $V^{\pi}(s)$ as the expected cumulative discounted reward associated with following this policy:
\begin{equation}
V^{\pi}(s) = \mathbb{E}_{\pi, \mathcal{P}}\left[ \sum_{t=0}^{T} \gamma^t r_{t+1} | s_0 = s \right]\notag
\end{equation}
In RL, we are interested in finding an optimal policy $\pi^*$ that results in the maximum value function 
$\pi^* \in \arg \max_{\pi} V^{*}$, where $V^{*}(s) \geq V^{\pi}(s), \forall \pi,s$.

% A Partially Observable Markov Decision Process (POMDP) is a generalization of the previously defined MDPs, where the agent cannot observe the full/direct state of the underlying MDP, but only has access to some observations $o$ generated via a observation function $\mathcal{O}(o| s')$. Given this observation and without having access to the true state of the MDP $s'$, to do inference and planning in this setup, the agent has to maintain a belief over this underlying state $s'$:
% \begin{equation}
%     b(s') = P(s'|o') \propto \mathcal{O}(o'|s') \int_{\mathcal{S}} \left[\mathcal{P}(s'|s,a) b(s) ds \right]
% \end{equation}

\subsection{MDP with another agent}

% Consider a base MDP with only the learning agent $M_l = \langle \mathcal{S}_l, \mathcal{A}_l, \mathcal{P}_l, \mathcal{R}, \gamma \rangle$, an expert MDP with only the expert agent $M_e = \langle \mathcal{S}_e, \mathcal{A}_e, \mathcal{P}_e, \mathcal{R}, \gamma \rangle$ and an expert policy $\pi_e$. 

By introducing the expert (following policy $\pi_e$) in the learner's environment and making it  visible in the observational state of the learner, we change the learner's MDP. The resulting MDP can be parameterized as follows: $\tilde {\mathcal{M}} = \langle \tilde{\mathcal{S}}, \mathcal{A}, \tilde{\mathcal{P}}, \mathcal{R}, \gamma \rangle$, where now the state space consists of: i) a part of the state space that can be directly influenced by the learner, we will refer to this part of the state space as the \textit{controllable} part of the state space $\mathcal{S}_c$ and ii) a part of state space that the learner does not have any direct control over, but this is still part of its observational state and includes useful information, $\mathcal{S}_{\lnot c}$. In our case, $\mathcal{S}_{\lnot c}$ will include observations corresponding to the presence of the teacher. Given this factorization, $\tilde{s} = (s_c, s_{\lnot c}) \in \mathcal{S}_c\times\mathcal{S}_{\lnot c}$ and the transition dynamics naturally factorizes over these dimensions:
%\begin{eqnarray}
     %\tilde{P}( (s'_l,s'_e) | (s_a, s_e), a_l) 
     %&=& \sum_{a_e} P( (s'_a,s'_e), a_e | (s_a, s_e), a) \\
    % &=& \sum_{a_e} P( (s'_a,s'_e)| (s_a, s_e), a, a_e) \pi_e(a_e|s_e) \\
   %  &=& \sum_{a_e} P(s'_a| s_a, a) P(s'_e| s_e, a_e) \pi_e(a_e|s_e) \\
  %   &=& \mathbb{E}_{a_e \sim \pi_e(.|s_e)} \left[ P(s'_a| s_a, a) P(s'_e| s_e, a_e) \right]\\
 %    &=& P(s'_a| s_a, a) P^{\pi_e}(s'_e| s_e)
%\end{eqnarray}
\begin{equation*}
\tilde{\mathcal{P}}( (s'_c,s'_{\lnot c}) | (s_c, s_{\lnot c}), a)= \mathcal{P}(s'_{c}| s_{c}, a)\mathcal{P}(s'_{\lnot c}|s_{\lnot c},\pi_e) = \mathcal{P}(s'_{c}| s_{c}, a)\mathcal{P}^{\pi_e}(s'_{\lnot c}|s_{\lnot c}).    
\end{equation*}

Thus, if the policy of the expert (other agent) is stationary, introducing the extra agent results in another valid MDP. On the other hand, if the policy of the other agent in not stationary, the induced transition kernel will change every time there is a switch in the expert's policy.

\subsection{Formal setup}
We consider a set of MDPs that share states, actions, transition dynamics and discount factor, but differ in the reward function $ G = \{ \mathcal{M}_t | \mathcal{M}_t = \langle \mathcal{S}, \mathcal{A}, \mathcal{P}, \mathcal{R}_t, \gamma \rangle \}$. Now, let us consider uniformly sampling one of these MDPs $\mathcal{M}_t \sim U(G)$ and unrolling one episode given this sampled choice. Once the episode has terminated, we re-sample from $G$ and repeat this process at the start of each episode. This procedure defines another MDP
$\mathcal{M} = \langle \mathcal{S}, \mathcal{A}, \mathcal{P}, \mathcal{R}, \gamma \rangle $ where $\mathcal{R} = \mathbb{E}_{\mathcal{M}_t} \left[\mathcal{R}_t\right]$\footnote{Note: 
\begin{eqnarray*}
     \mathbb{E}_{\mathcal{M}_t} [V_{\mathcal{M}_t}^{\pi}] 
     = \mathbb{E}_{\mathcal{M}_t} \left[\mathcal{R}_t(s,a) + \gamma \mathbb{E}_{s' \sim P(.|s,a)}V_{\mathcal{M}_t}^{\pi}(s')\right]
     =\mathbb{E}_{\mathcal{M}_t} \left[\mathcal{R}_t(s,a)\right] + \gamma \mathbb{E}_{s' \sim P(.|s,a)} [V_{\mathcal{M}_t}^{\pi}(s')] 
\end{eqnarray*}\vspace{-3mm}}. 
This holds only when the transitional dynamics $\mathcal{P}$ is shared across the candidate MDPs.
We are interested in the policy $\pi$ that performs well, in expectation across this family of MDPs. If the teacher is not part of the environment, $\mathcal{S}$ is isomorphic with $\mathcal{S}_{c}$, leading to:
\begin{equation}
    \pi \in \arg\max_{\pi} \left( \mathbb{E}_{\mathcal{M}_t \sim U(G)} [V_{\mathcal{M}_t}^{\pi}]\right)
    \label{eq:default_stationary_policy}
\end{equation}
In the following we will distinguish the MDPs with the teacher as $\mathcal{M}$ and the one including the expert in the same environment $\tilde{\mathcal{M}}$ and the respective optimal policies $\pi^*$ and $\tilde{\pi^*}$.

\textbf{Introducing an expert into the environment}. The question we would like to answer is whether it is possible to do better than the above stationary policy, when placing the student in the augmented MDP $\tilde{\mathcal{M}}$. Now our agent has access to another agent's behaviour, but \textit{only through its own observational state}. Another consequence of moving to $\tilde{\mathcal{M}}$ is an augmentation in the state space, which results also in an expansion of the policy space under consideration. Since the set of possible policies in $\tilde{\mathcal{M}}$ is a superset of the policies in $\mathcal{M}$, it is easy to see that the optimal policy in the augmented space is at least as good as the optimal policy in the original MDP $\mathcal{M}$:
\begin{eqnarray*}
    \mathbb{E}_{\mathcal{M}_t} [V_{\mathcal{M}_t}^{\tilde{\pi}}] 
    &=& \mathbb{E}_{\mathcal{M}_t} \left[R_t(s,a)\right] + \gamma  \mathbb{E}_{\mathcal{M}_t} \mathbb{E}_{\tilde{s}' \sim P_t(.|\tilde{s},a)} [V_{\mathcal{M}_t}^{\tilde{\pi}}(\tilde{s}')] \\
    &=& \mathbb{E}_{\mathcal{M}_t} \left[R_t(s_c,a)\right] + \gamma  \mathbb{E}_{\mathcal{M}_t} \mathbb{E}_{{s}'_c \sim P(s'_c|s_c,a), s'_{\lnot c} \sim P^{\pi_e}(s'_{\lnot c}|s_{\lnot c})} [V_{\mathcal{M}_t}^{\tilde{\pi}}(s'_c,s'_{\lnot c})] \\
    &\leq& \mathbb{E}_{\mathcal{M}_t} \left[R_t(s_c,a)\right] + \gamma  \mathbb{E}_{\mathcal{M}_t} \mathbb{E}_{{s}'_c \sim P(.|{s_c},a)} [V_{\mathcal{M}_t}^{\pi}({s}'_c)] 
\end{eqnarray*}
The agent can leverage the behaviour of the teacher in order to learn a policy $\pi(a|s_c,s_{\lnot c})$ that is better than $\pi(a|s_c)$ defined in eq. \ref{eq:default_stationary_policy}.
Given this setup, let us take a closer look at two particular case studies, where this might occur:

i) \textbf{Imitation}. Let us consider the case where $\mathcal{A}=\mathcal{A}_e$, where $\mathcal{A}_e$ is the action-space of the expert.
Let us assume the teacher's policy is better than the stationary policy defined in eq. \ref{eq:default_stationary_policy}, $\pi^*_{\mathcal{M}}$ -- otherwise there is no incentive to deviate from $\pi^*_{\mathcal{M}}$ if the behaviour proposed for cloning is inferior. If $s_{\lnot c}$ includes the actions taken by the expert, then imitation can trivially take place, by just reading off the desired action from $s_{\lnot c}$ : $\pi(a|s_c, s_{\lnot c}) := \pi (a_e | s_c, s_{\lnot c})$\footnote{We assume the two agents have the same starting position, but this is can relaxed, by first employing a policy that gets the learning agent close enough to the teacher, after which the mimicking behaviour can occur. We assume the learner to always be one step behind the teacher, but longer time difference can be negotiated via a memory component.}. It is actually a known result that observing experts' actions is mandatory to achieve actual imitation~\cite{ho2016generative}. In a purely observational setting though, the student would not have typically access directly to the actions performed by the expert. Nevertheless, we also know that in a deterministic environment, observing the effects of one's actions in the environment is enough to infer/encode the action that was performed. To learn this mapping we need to 'remember' or keep track of at least two consecutive observations of the teacher's state. Thus, if imitation is to emerge, it can only do so in a system able to distill this information from the observational state. For this our agents will need memory or augmentation of the state space to include several time steps.

ii) \textbf{Information seeking behaviour}. If the MDP ($\mathcal{M}_t$) is identified, then the learner's policy can become independent of the teacher as the task can be optimally performed by the learning agent, without any input from the teacher.
We denote $\pi^*_{\mathcal{M}_t}$ the optimal policy in $\mathcal{M}_t$. We will make the additional assumption that given $t$ the optimal policy in the MDP with or without the expert is the same. Formally, $\pi^*_{\mathcal{M}_t} = \pi^*_{\tilde{\mathcal{M}}_t}$. Thus, if the identity $t$ of the task is known, the optimal behaviour of the agent would be to just switch between these optimal policies given the context $t$: $\tilde{\pi}(a|\tilde{s}, t) = \pi^*_{\mathcal{M}_t} (a|s_c) $. This policy is optimal for each of the sampled MDPs and results in an optimal behaviour in $\tilde{\mathcal{M}}$, provided the context $t$ is known.
If this information can be distilled from the observation of the other expert, the learner can improve over the stationary policy defined in eq. \ref{eq:default_stationary_policy}. Thus if $\exists g: \mathcal{S}_l \rightarrow \mathbb{N}_T $ s.t. $t = g(s_{\lnot c})$, then $\exists$ a stationary policy in the augmented state space $\tilde{\mathcal{S}}$ that (can) outperfom $\pi^*_{\mathcal{M}}$:
\begin{equation*}
    \tilde{\pi}(a|\tilde{s})  = \tilde{\pi}(a|\tilde{s}, g(s_{\lnot c})) = \pi^*_{\mathcal{M}_t} (a|s_c) 
\end{equation*}
Note that $g$ can, in principle, take into account in its computation, a series of observations of the expert (several steps of its trajectory). This can be practically implemented by stacking several recent observations in our current state space, or relying on a memory/recurrent model to distill this temporal information as part of its hidden state and infer the current context.

\section{Experiments}
\label{sec:exp}

For our experiments, we choose a widely used DeepRL algorithm: the Asynchronous Advantage Actor-Critic (A3C) algorithm \cite{mnih2016asynchronous}. This learns both a policy (the actor), $\pi_{\theta_{\pi}}(a_t|s_t)$  and value function (the critic)
 $V_{\theta_V} (s_t)$ given a state observation $s_t$. The two approximations share the intermediate representation and only diverge in the final fully-connected layer. The policy is given by a softmax over actions. 
The setup closely follows \cite{mnih2016asynchronous} including the entropy regularization and the addition of an LSTM layer to incorporate memory. We use both a simple feed-forward and a recurrent version in our experiments. For further details, we refer the reader to the original paper.

\begin{figure}
  \centering
  \subfloat[Level 1]{\includegraphics[width=0.3\textwidth]{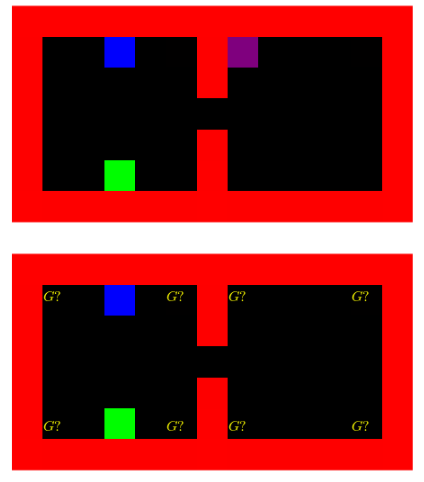}\label{fig:level1}}
  \hfill
  \subfloat[Level 2]{\includegraphics[width=0.3\textwidth]{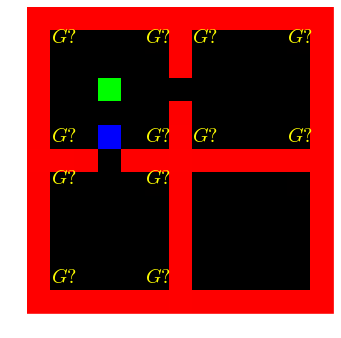}\label{fig:level2}}
  \hfill
  \subfloat[Level 3]{\includegraphics[width=0.3\textwidth]{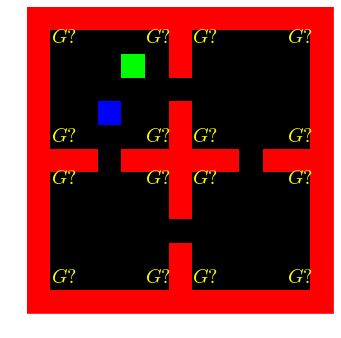}\label{fig:level3}}
  \caption{Environment snapshots}
    \vspace{-3mm}
\end{figure}

As a set of scenarios, we consider a series of simple navigation tasks. 
We first start with a two-room layout (Figure \ref{fig:level1}) and place the two agents in this environment. Possible goal locations considered are the corners of each of the two rooms (8 in total). At the start of each episode, we sample uniformly the location of the goal and make this information available to the teacher at all times. For the learning agent, we consider both including and occluding the goal in its observational space.
In the second version of this task, the student does not know which goal is activated at this point in time. It can only learn that the possible positions of the reward are the eight corners. If it ignores the expert's behaviour the optimal stationary policy would be to visit all of these locations in the minimum time possible. 
%Since the goals are place at the corners of the rooms the optimal 'blind' strategy is to go around the walls 'touring' each room. 
On the other hand, the teacher has access to the goal and can seek it directly. By observing its behaviour, the learner could potentially disentangle which corner is active, or at least narrow down the correct room, and use this information to improve its performance. 

\subsection{Global view: task identity occluded, but perfect information otherwise.}
\begin{figure}
  \centering
  \subfloat{\includegraphics[width=0.5\textwidth]{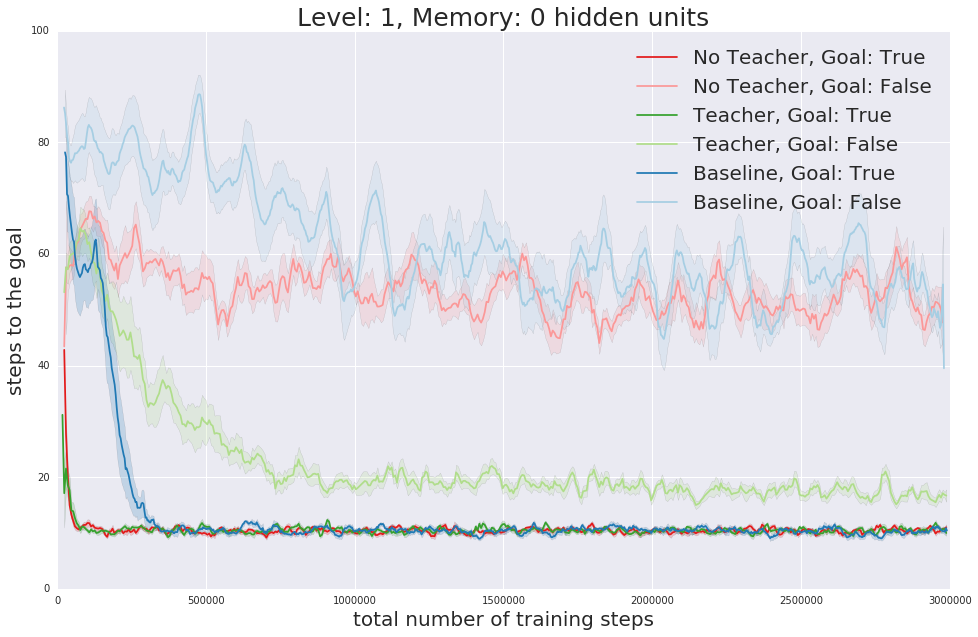}\label{fig:global_level1_memory0}}
  \hfill
  \subfloat{\includegraphics[width=0.5\textwidth]{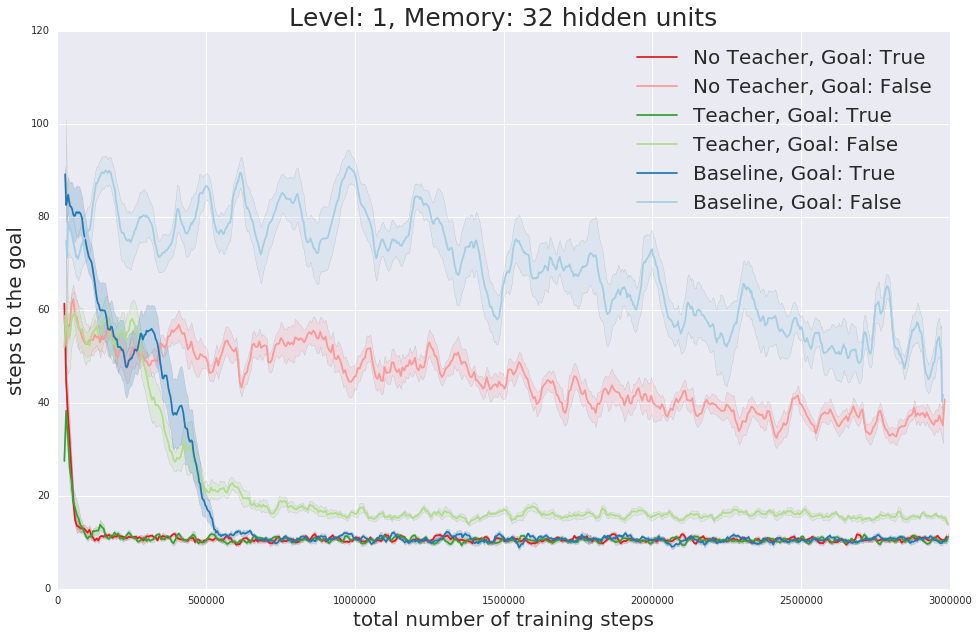}\label{fig:global_level1_memory32}}
  \caption{Level 1: Performance during training as measured by the number of steps to the goal. Red curves: learning agent alone in the environment. Green curves: learning agent shares the environment with a teacher. Blue curves: learning from scratch. Bold coloured curves: the goal is present (LAGT, LAG). Light coloured curves: the goal is occluded (LA, LAT).}
  \vspace{-5mm}
\end{figure}
We provide the student with a top-view representation of the environment. We use a one-hot encoding for each element present in the environment: one channel for the layout (L), one for the position of the agent (A), one for the position of the teacher (T), one for the position of the goal (G). We implement four variations of this observational space: LA, LAG, LAT, LAGT.

We run this experiment with an A3C agent with two small convolutional layers, followed by a fully connected layer. The results are displayed in Figures \ref{fig:global_level1_memory0} (Feed-forward network), \ref{fig:global_level1_memory32} (LSTM with 32 units). The first thing to notice is that, with perfect information, the learned policies with or without the teacher have the same asymptotic performance. Moreover the speed at which this is reached is not negatively impacted by the expansion of the state space. Thus, for this experiment we conclude that the presence or absence of the teacher does not have an impact on the learning process -- this is observed across multiple experiments (Appendix A).

In the second case, when the student has impoverished knowledge of the environment, we can clearly spot a difference between the performance of the student when sharing in the same environment with the teacher and acting on its own. For the first part of the training, the agent achieves similar performance, but at some point, the agent sharing the same environment with the teacher manages to leverage this extra signal to significantly improve its policy. 

\begin{figure}
  \centering
  \subfloat{\includegraphics[width=0.5\textwidth]{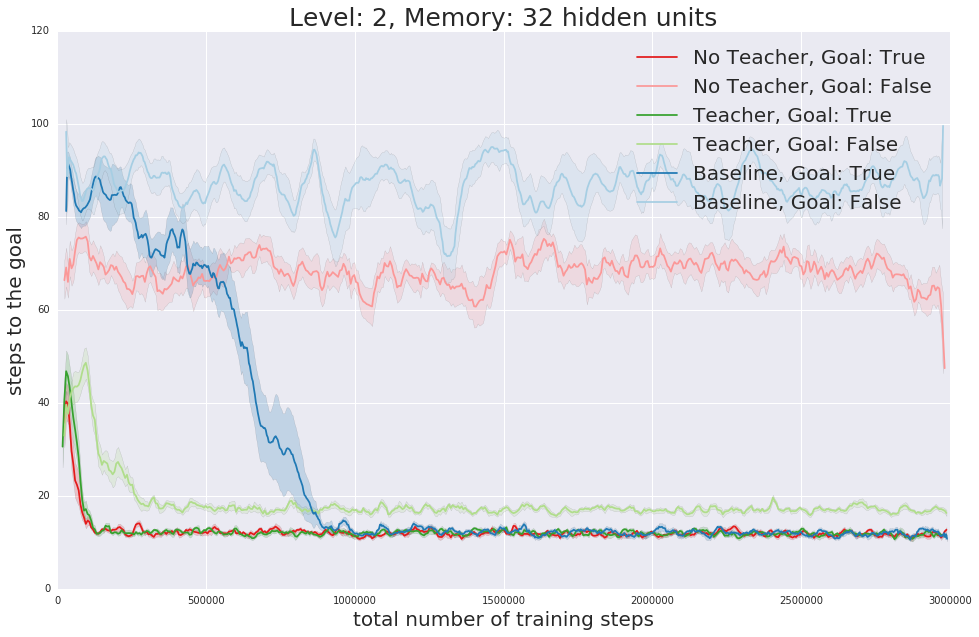}\label{fig:global_level2_memory32}}
  \hfill
  \subfloat{\includegraphics[width=0.5\textwidth]{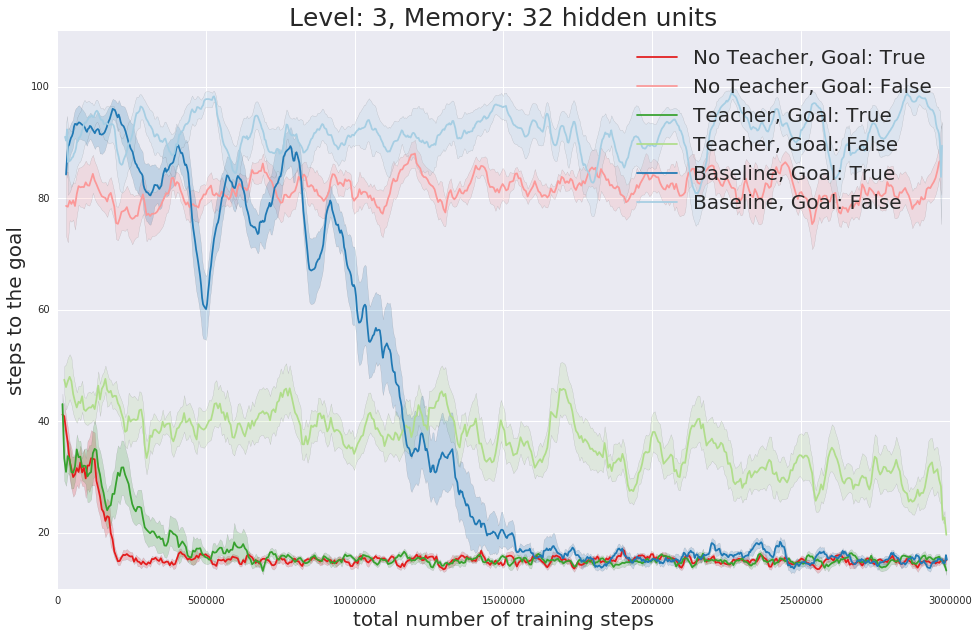}\label{fig:global_level3_memory32}}
    \caption{Level 2 and 3: Performance during training as measured by the number of steps to the goal. Red curves: learning agent alone in the environment. Green curves: learning agent shares the environment with a teacher. Blue curves: learning from scratch (w/o curriculum). Bold coloured curves: the goal is present (LAGT, LAG). Light coloured curves: the goal is occluded (LA, LAT). When the goal is occluded we can see that the agent can do better by leveraging the expert.}
  \vspace{-5mm}
 \end{figure}
As we observed that the student is able to exploit the expert's behaviour for its own benefit, we now want to know if it is a transferable knowledge across different environments. To test this out, we constructed two other ``levels'', by adding an additional room each time and thus smoothly increasing the difficulty of the task. A depiction of these levels can be found in Figure \ref{fig:level2}, \ref{fig:level3}. When extending the environment, we also naturally extend the number of possible goal locations. 
%Thus the complexity of the task increases with each level. 
The average number of steps to the goal increases and when the goal location is occluded the 'blind' stationary policy will be quite expensive, thus it becomes more and more crucial for the agent to leverage the expert.

We train on these new levels in a curriculum fashion: first we train on level 1, then continue training on level 2, and finally on level 3. This is mainly done to speed up training time, but also we expect the agent to learn the importance of the expert in level 1 and continue to employ this knowledge in its learning process in the next levels.
The specification of the observational state of the learner is maintained throughout the curriculum: with/without the teacher, with/without the goal visible. The results are compiled in Figure \ref{fig:global_level2_memory32}, \ref{fig:global_level3_memory32}. For reference, we included the baseline of learning in these levels starting from a random initialization, with the curriculum. The curriculum helps the learning process both when the goal is visible and when it is occluded: the performance is slightly better to begin with and then convergence is achieved considerably faster. Most importantly, the presence of the expert consistently improves the policy of the 'blind' agent across all levels. It is worth noting also, that for the last level when the goal is occluded, both the baseline and the lone agent are not always able to complete the task, whereas the agent leveraging the teacher can almost instantaneously complete the task at each episode (the transfer to level 3 is almost zero-shot).
\subsection{Local view: agent needs to actively keep track of the expert, if useful}
We have seen in the previous section that by just using RL on the same reward signal, we can obtain different policies when environment is augmented with the presence of an expert. Although occluding the goal might seem like a somewhat artificial task, this is just a proxy for impoverished or imperfect information. Yet we aknowledge that a more natural example of this is partial observability -- when the agent has only a local view of the environment. We thus simulate this in our environment by taking a local window of observation centered around the agent. The rest of the setup remains the same, but note that we are changing the nature of the problem quite drastically. This is especially true if one thinks about the setting in which the teacher is in the picture. In this new scenario, the learner will have to actively pursue the expert if it has any hope of benefiting from its presence. If it cannot see the expert, the learner cannot learn from it. %This is different from the previous setting where position of the teacher was always visible to the LAT, LAGT agents.
\begin{figure}
  \centering
  \subfloat{\includegraphics[width=0.5\textwidth]{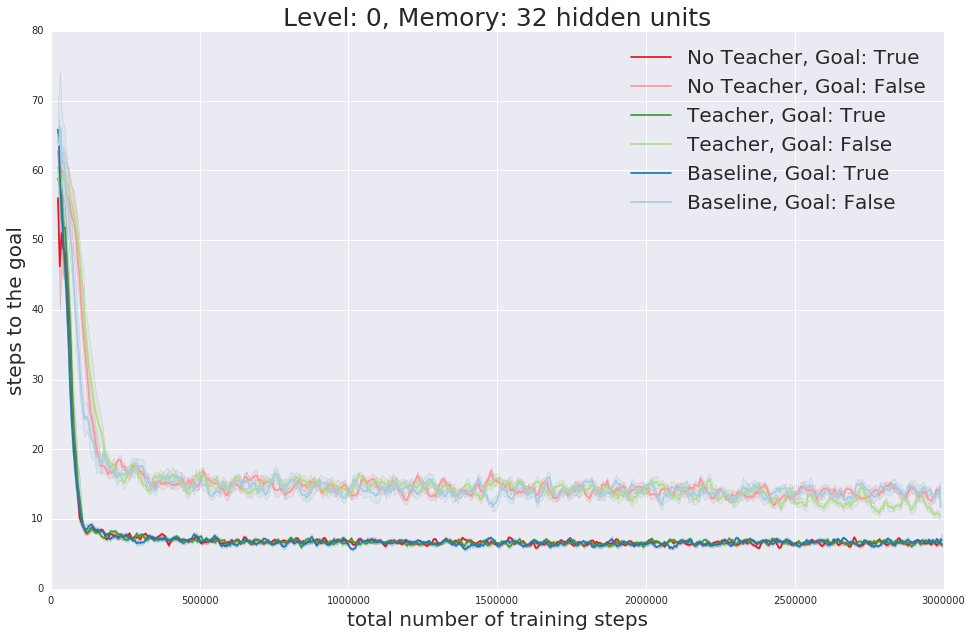}}
  \hfill
  \subfloat{\includegraphics[width=0.5\textwidth]{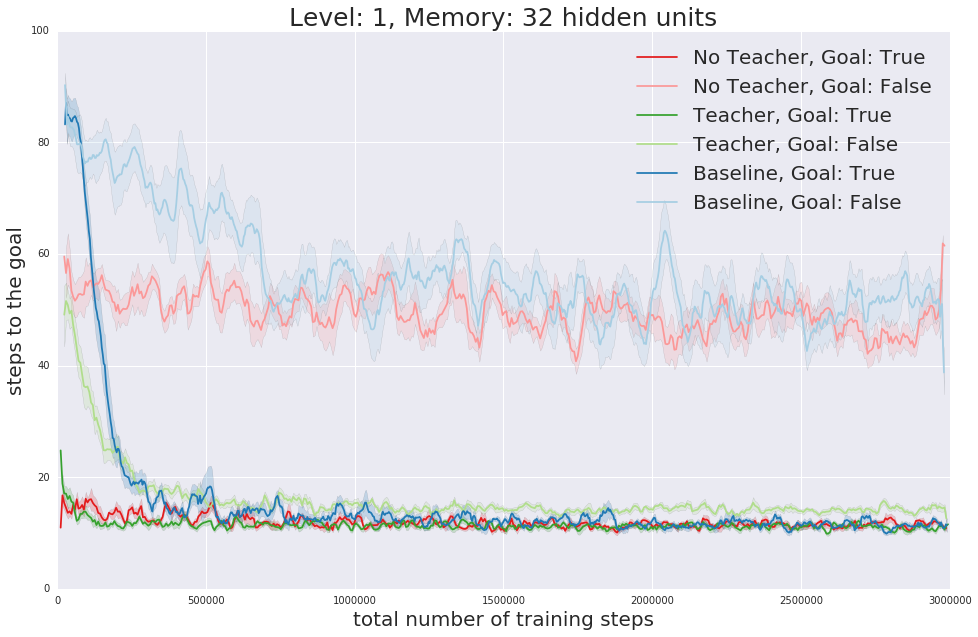}}
  \hfill
  \subfloat{\includegraphics[width=0.5\textwidth]{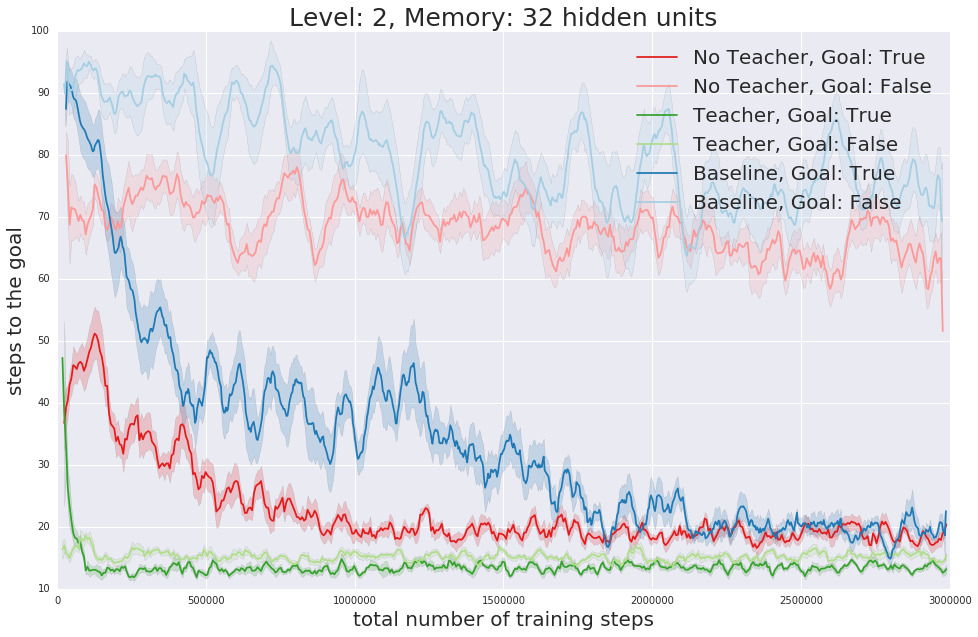}}
  \hfill
  \subfloat{\includegraphics[width=0.5\textwidth]{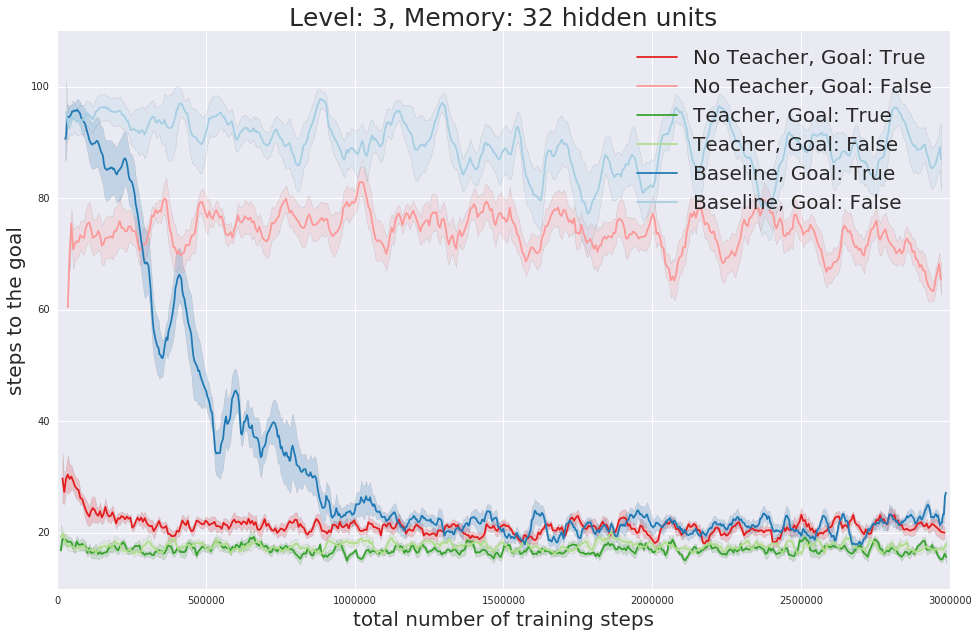}}
  \caption{Local View:  Performance during training (average steps to the goal), in all levels of the curriculum. Green curves: the learning agent shares the environment with the teacher. Blue curves: learning from scratch in this environment. Bold coloured curves: the goal is present (LAGT, LAG). Light coloured curves: the goal is occluded (LA, LAT). The teacher's presence has a significant impact on the performance in the training and quality of the end policy. It universally improving, or at least matching the performance of the lone agent (level 0). }
  \label{fig:local_view_results}
  \vspace{-5mm}
\end{figure}

Because of the increased complexity, we now start with only one of the two rooms in level 1 and we basically treat this as pre-training of simple navigation skills with local view. In this level, the learner can learn to navigate to the goal and visit the potential locations where the goal is hidden, but we do not observe any benefit from having the expert present. The size of the local window of observation is the size of the room -- if the agent is in the middle of the room it can see the whole room. This always means that the problem of keeping track of the other agent is not very difficult in this first setting. The size of the window ensures that the expert is seen in a good proportion of the observations. The learning curves can be found in Figure \ref{fig:local_view_results}. 
Nevertheless, a different story emerges when we look at the training in level 1. First, we observe that when the goal is hidden, but the teacher is present, we get an improvement in policy over the scenario where the agent is alone in the environment. This is consistent with what we have seen before in the global view. When the potential benefit over the 'blind' stationary policy is big enough, the student begins to 'pay' attention to the teacher and uses these observations to improve its own policy. This carries on to levels 2 and 3. Furthermore, at these last levels, we can see that in the cases where the teacher is present, the asymptotic performance of the agent matches or slightly outperforms that of the lone agent with the \textit{goal visible}. This is remarkable as we are now seeing an improvement in policy even when the goal is (locally) visible. This is because in partial observability the trajectory of the expert still contains valuable information about where the goal might be. At each step, the teacher narrows down the possible locations. 

\subsection{Breaking away from the teacher}
The fact that the final performance is independent of the goal's presence or absence in the state space suggests that the only information the agent learns to exploit is the behaviour of the teacher. Visual inspection of the final policies fully supports this intuition. The behaviour that emerges is the following: the agent seeks the teacher and tries to get as close as possible to it, after which it simply follows it to the goal.\footnote{
Videos of these behaviours can be found in the Supplementary Material and further illustrations \href{https://www.youtube.com/playlist?list=PL9d1eKcBwMR9_Aw0PgAZZIQPlYPLhxscI}{\underline{here}}.} This is potentially a very useful, transferable strategy to a different environment. To test this idea, we expand further our environment to 9 rooms and without any further learning, the previously trained agent can successfully negotiate this new environment with the help of the teacher and succeeds in finding goals it has never seen before.
Nevertheless, relying always on the teacher is somewhat unsatisfactory. In fact, if we do eliminate the teacher, the learning agent is quite lost. It will only reach for the goals currently visible if any, and otherwise will continue waiting for the other agent to appear, not even attempting to leave the spawning room. Yet, this is an agent that has previously negotiated this environment with the help of the agent, but has not retained the particularities of this environment because it did not need to. 
%The only thing it ever had to worry about is keeping track of the other agent. 
%This is a much smaller local problem that our agent needs to solve at each step, but it never had to worry about the global problem, which is implicitly solved by the teacher.
In order to break this dependence, we propose a simple (curriculum) strategy: we mask the presence of the teacher with some probability that increases over time. When this probability reaches 1 and the agent becomes completely independent of the teacher. We start with a masking probability of $0.25$, then after 250k steps ($1/4$ of our normal training period for each level) we decrease it to $0.5$, then $0.75$ and finally $1.0$. At the end of this process we successfully achieved an agent that can tackle the environment on its own, without the teacher.

% \subsection{Suboptimal teacher}
% Teacher behaves suboptimally, can the student do better on its own? Does it pick up on that strategy? Abandoning the suboptimal player. 
% Encourage this behaviour by penalizing actions or having a restrictive action budget (this might emerge naturally by the environment cap over frames.).

% \subsection{[Next week] Observing consequence of teacher's behaviour and adjust accordingly}

% \subsection{Discussion}

% Adding the teacher we are expanding considerably the state space of the learning agent, thus on top of the RL policy, the student now also need to learn to ignore this extra signal in the state space. What we have noticed in practice is that the presence of the other agent does not have a negative influence on the learning process or the emerging policy. This might be because we are not really increasing the complexity of the policy being learnt, but just adding some complexity to the representation -- and our tasks are fairly simple on this end, but it is a valid concern in more complex domain. 

% Importance of curriculum

% Importance of memory

\section{Conclusion and further work}
\label{sec:ccl}
This paper illustrates a new case of meta learning through reinforcement learning. It demonstrates that observational learning can emerge from reinforcement learning, combined with memory. We argued that observational learning is an important learning mechanism that can be a 'cheaper' alternative than some of the current approaches to learning from experts. In particular, this approach does not require any explicit modelling of the expert, nor mapping their experience in our agent's state and action space. 
We have shown in our experiments, that relying only of the (sparse) reward signal given by the environment coupled with only the sheer presence of another agent, can lead to a variety of behaviours, ranging smoothly between imitation and information seeking. We also demonstrated that, via a curriculum, we could end up with an autonomous agent that solves tasks without the presence of the teacher although it learned through observations of a teacher.

This is an initial work which should be extended further, in other settings. Especially, we want to test scenarios where the goal of the teacher and of the learner are not strictly aligned. We performed a preliminary study where there is a negative correlation (\textit{e.g.} they are in opposite directions in the room). Results are provided in Appendix. In the same way, we should study how optimal the teacher has to be, how much its motivation should be correlated to the one of the student.


% (Don't include this in the submission)

%% file: ms.bbl
\begin{thebibliography}{10}

\bibitem{sutton1998reinforcement}
Richard~S Sutton and Andrew~G Barto.
\newblock {\em Reinforcement learning: An introduction}.
\newblock Cambridge Univ Press, 1998.

\bibitem{mnih2015human}
Volodymyr Mnih, Koray Kavukcuoglu, David Silver, Andrei~A Rusu, Joel Veness,
  Marc~G Bellemare, Alex Graves, Martin Riedmiller, Andreas~K Fidjeland, Georg
  Ostrovski, et~al.
\newblock Human-level control through deep reinforcement learning.
\newblock {\em Nature}, 518(7540):529--533, 2015.

\bibitem{lecun2015deep}
Yann LeCun, Yoshua Bengio, and Geoffrey Hinton.
\newblock Deep learning.
\newblock {\em Nature}, 521(7553):436--444, 2015.

\bibitem{hochreiter1997long}
Sepp Hochreiter and J{\"u}rgen Schmidhuber.
\newblock Long short-term memory.
\newblock {\em Neural computation}, 9(8):1735--1780, 1997.

\bibitem{chung2015gated}
Junyoung Chung, Caglar Gulcehre, Kyunghyun Cho, and Yoshua Bengio.
\newblock Gated feedback recurrent neural networks.
\newblock In {\em International Conference on Machine Learning (ICML)}, 2015.

\bibitem{mnih2016asynchronous}
Volodymyr Mnih, Adria~Puigdomenech Badia, Mehdi Mirza, Alex Graves, Timothy
  Lillicrap, Tim Harley, David Silver, and Koray Kavukcuoglu.
\newblock Asynchronous methods for deep reinforcement learning.
\newblock In {\em International Conference on Machine Learning (ICML)}, 2016.

\bibitem{mirowski2016learning}
Piotr Mirowski, Razvan Pascanu, Fabio Viola, Hubert Soyer, Andy Ballard, Andrea
  Banino, Misha Denil, Ross Goroshin, Laurent Sifre, Koray Kavukcuoglu, et~al.
\newblock Learning to navigate in complex environments.
\newblock In {\em Proceedings of the International Conference on Learning
  Representations (ICLR)}, 2017.

\bibitem{levine2016end}
Sergey Levine, Chelsea Finn, Trevor Darrell, and Pieter Abbeel.
\newblock End-to-end training of deep visuomotor policies.
\newblock {\em The Journal of Machine Learning Research}, 17(1):1334--1373,
  2016.

\bibitem{bandura1977social}
Albert Bandura and Richard~H Walters.
\newblock {\em Social learning theory}.
\newblock Prentice-Hall Englewood Cliffs, NJ, 1977.

\bibitem{bandura1963social}
Albert Bandura and Richard~H Walters.
\newblock {\em Social learning and personality development}, volume~14.
\newblock JSTOR, 1963.

\bibitem{heyes1993imitation}
Cecilia~M Heyes.
\newblock Imitation, culture and cognition.
\newblock {\em Animal Behaviour}, 46(5):999--1010, 1993.

\bibitem{schaal1999imitation}
Stefan Schaal.
\newblock Is imitation learning the route to humanoid robots?
\newblock {\em Trends in cognitive sciences}, 3(6):233--242, 1999.

\bibitem{argall2009survey}
Brenna. Argall, Sonnia Chernova, Manuella Veloso, and Brett Browning.
\newblock A survey of robot learning from demonstration.
\newblock {\em Robotics and autonomous systems}, 57(5):469--483, 2009.

\bibitem{pomerleau1989alvinn}
Dean~A. Pomerleau.
\newblock Alvinn: An autonomous land vehicle in a neural network.
\newblock Technical report, DTIC Document, 1989.

\bibitem{ratliff2007imitation}
Nathan Ratliff, J.Andrew Bagnell, and Siddhartha~S. Srinivasa.
\newblock Imitation learning for locomotion and manipulation.
\newblock In {\em Proceedings of the IEEE-RAS International Conference on
  Humanoid Robots}, 2007.

\bibitem{russell1998learning}
S.~Russell.
\newblock Learning agents for uncertain environments.
\newblock In {\em Proceedings of the Conference on Learning Theory (COLT)},
  1998.

\bibitem{ng2000algorithms}
Andrew Ng and Stuart Russell.
\newblock Algorithms for inverse reinforcement learning.
\newblock In {\em Proceedings of the International Conference on Machine
  Learning (ICML)}, 2000.

\bibitem{atkeson1997robot}
C.G. Atkeson and S.~Schaal.
\newblock Robot learning from demonstration.
\newblock In {\em Proceedings of the International Conference on Machine
  Learning (ICML)}, 1997.

\bibitem{ziebart2008maximum}
Brian~D Ziebart, Andrew~L Maas, J~Andrew Bagnell, and Anind~K Dey.
\newblock Maximum entropy inverse reinforcement learning.
\newblock In {\em Proceedings of the Annual Meeting or AAAI}, 2008.

\bibitem{kolter2008hierarchical}
J~Zico Kolter, Pieter Abbeel, and Andrew~Y Ng.
\newblock Hierarchical apprenticeship learning with application to quadruped
  locomotion.
\newblock In {\em Advances in Neural Information Processing Systems (NIPS)},
  2008.

\bibitem{ross2011reduction}
St\'ephane Ross, Geoffrey~J. Gordon, and J.~Andrew Bagnell.
\newblock A reduction of imitation learning and structured prediction to
  no-regret online learning.
\newblock In {\em Proceedings of the International Conference on AI and
  Statistics (AISTATS)}, 2011.

\bibitem{ho2016showing}
Mark~K Ho, Michael Littman, James MacGlashan, Fiery Cushman, and Joseph~L
  Austerweil.
\newblock Showing versus doing: Teaching by demonstration.
\newblock In {\em Advances in Neural Information Processing Systems (NIPS)},
  2016.

\bibitem{ho2016generative}
Jonathan Ho and Stefano Ermon.
\newblock Generative adversarial imitation learning.
\newblock In {\em Advances in Neural Information Processing Systems (NIPS)},
  2016.

\bibitem{piot2014predicting}
Bilal Piot, Olivier Pietquin, and Matthieu Geist.
\newblock Predicting when to laugh with structured classification.
\newblock In {\em Proceedings of INTERSPEECH}, 2014.

\bibitem{neu2009training}
Gergely Neu and Csaba Szepesv{\'a}ri.
\newblock Training parsers by inverse reinforcement learning.
\newblock {\em Machine learning}, 77(2), 2009.

\bibitem{stadie2017third}
Bradly~C Stadie, Pieter Abbeel, and Ilya Sutskever.
\newblock Third-person imitation learning.
\newblock In {\em Proceedings of the International Conference on Learning
  Representations (ICLR)}, 2017.

\bibitem{abbeel2004apprenticeship}
Peter Abbeel and Andrew~Y. Ng.
\newblock Apprenticeship learning via inverse reinforcement learning.
\newblock In {\em Proceedings of the International Conference on Machine
  Learning (ICML)}, 2004.

\bibitem{piot2014boosted}
Bilal Piot, Matthieu Geist, and Olivier Pietquin.
\newblock Boosted and reward-regularized classification for apprenticeship
  learning.
\newblock In {\em Proceedings of the 2014 international conference on
  Autonomous agents and multi-agent systems (AAMAS)}, 2014.

\bibitem{piotECML}
Bilal Piot, Matthieu Geist, and Olivier Pietquin.
\newblock Learning from demonstrations: is it worth estimating a reward
  function?
\newblock In {\em Proceedings of the European Conference on Machine Learning
  (ECML)}, 2013.

\end{thebibliography}
